\documentclass{article}

\usepackage{arxiv}

\usepackage[utf8]{inputenc} % allow utf-8 input
\usepackage[T1]{fontenc}    % use 8-bit T1 fonts
\usepackage{hyperref}       % hyperlinks
\usepackage{url}            % simple URL typesetting
\usepackage{booktabs}       % professional-quality tables
\usepackage{amsfonts}       % blackboard math symbols
\usepackage{nicefrac}       % compact symbols for 1/2, etc.
\usepackage{microtype}      % microtypography
\usepackage{lipsum}
\usepackage{graphicx}
\usepackage{bm}
\usepackage{amsfonts,amssymb} 
\usepackage{footnote}

\graphicspath{ {./Graph/} }

\title{Meta-Embeddings Based On Self-Attention}

\author{
 Qichen Li\\
 Massachusetts Institute of Technology \\
 \texttt{liqichen@mit.edu}
   \And
 Yuanqing Lin \\
 Peking University \\
 \texttt{yuanchinglin@pku.edu.cn}
  \And
 Luofeng Zhou \\
 Columbia University \\
 \texttt{luofeng.zhou@columbia.edu}
  \And
 Jian Li \\
 Tsinghua University \\
 \texttt{lijian83@mail.tsinghua.edu.cn}\\
}
\begin{document}
\maketitle

\begin{abstract}
Creating meta-embeddings for better performance in language modelling has received attention lately, and methods based on concatenation or merely calculating the arithmetic mean of more than one separately trained embeddings to perform meta-embeddings have shown to be beneficial. In this paper, we devise a new meta-embedding model based on the self-attention mechanism, namely the Duo. With less than 0.4M parameters, the Duo mechanism achieves state-of-the-art accuracy in text classification tasks such as 20NG. Additionally, we propose a new meta-embedding sequece-to-sequence model for machine translation, which to the best of our knowledge, is the first machine translation model based on more than one word-embedding. Furthermore, it has turned out that our model outperform the Transformer not only in terms of achieving a better result, but also a faster convergence on recognized benchmarks, such as the WMT 2014 English-to-French translation task. 
\end{abstract}

\section{Introduction}
Transformer  \cite{vaswani2017attention} , Recurrent neural networks with long short-term memory \cite{hochreiter1997long} and gated recurrent neural networks \cite{chung2014empirical}, have been firmly established as state of the art approaches in sequence modelling and machine translation \cite{sutskever2014sequence, bahdanau2014neural, cho2014learning}. Without one single exception, these models use the distributed vector representations of words, referred to as word embeddings, as their cornerstone. Furthermore, researches have shown that a word embedding set with better quality can benefit the whole model \cite{kocmi2017exploration}, and methods of "meta-embedding", first proposed by \cite{yin2016learning}, can yield an embedding set with improved quality. Therefore, meta-embedding can benefit the language modelling.

To yield an embedding set with better quality, several methods have been proposed in terms of meta-embeddings, e.g., 1\texttt{to}N+ \cite{yin2016learning} takes the ensemble of $K$ pre-trained embedding sets, and use a neural network to recover its corresponding vector within each source embedding set. An unsupervised approach is empolyed by \cite{bollegala2017think}: for each word, a representation as a linear combination of its nearest neighbours is learnt. Other methods, despite their simplicity, such as concatenation \cite{yin2016learning}, or averaging source word embedding \cite{coates2018frustratingly} has been used to provide a good baseline of performance for meta-embedding. In this work, we explore a new way of meta-embedding, namely, the Duo. To the best of our knowledge, our model is the first meta-embedding method based on self-attention mechanism. 

Our meta-embedding language learning model uses the Transformer as our back stone, which is a model architecture eschewing recurrence and relying only on attention, of which the mechanism is drawing of global dependencies between input and output. As recurrence is deducted, the parallelization is greatly enhanced. 
Because we use two times the embeddings to learn, the number of heads in the Transformer doubles and better performance is gained. Moreover, we use weight sharing in duo multi-head attention; thus the number of parameters is reduced.

The mechanism of the Duo is that instead of merely adding the dimension of word embedding, which leads to enormous increasing number in parameters, we use separately trained embedding as key and value for each word in the self-attention mechanism of the Transformer. As the number of word embedding doubles, the information in attention also doubles.  The discrepancy between the two pieces of independent embedding describes two different aspects of the same information. As our results demonstrate, this independence is quite beneficial to the training of the model.

Moreover, recent research \cite{tang2018self} has shown that the Transformer has shortcomings in long sequence learning, the Transformer-XL and other methods \cite{child2019generating, sukhbaatar2019adaptive} are therefore proposed to address the long sequence problem. The good news is that our model is very general that the Transformer-XL, along with other language models based on the Transformer, can employ the Duo mechanism to perform meta-embedding learning.

We examine our model in two representative tasks: the text classification task and the machine translation task. When it comes to a text classification problem, the Duo mechanism exploits the information in two pieces of word embedding, each separately trained, e.g., GloVe \cite{pennington2014glove} and fastText \cite{joulin2016bag}. This meta-embedding model allows the language model to have more previous knowledge in independent aspects, thus leading to a better result. The machine translation task is more tricky for meta-embedding learning, as it concerns the devising of a decoder. However, in this paper, we proposed a sequence-to-sequence meta-embedding language model to handle this problem, and the experiment shows that learning in such a way leads to better performance and a faster convergence.

All in all, the contributions of us are threefold:
\begin{itemize}
  \item We propose an attention-based way of meta-embedding for a better language modelling.
  \item To the best of our knowledge, we devise the first sequence-to-sequence encoder-decoder language model which directly uses two independent embeddings.
  \item The Duo mechanism we propose is very general and can be employed on any language model based on the Transformer.
  
\end{itemize}

\section{Background}
\label{sec:headings}
For the deep learning method in the text classification problem, word embedding has been a focus of much research \cite{mikolov2013distributed, pennington2014glove} as several studies showed that the text classification task depends enormously on the effectiveness of the work embedding \cite{shen2018baseline, wang2018joint}. The first part of our work focuses on combining different pre-trained word embedding for text classification. In other words, the Duo mechanism enables two different pieces of pre-trained word embedding to perform on the same stage.

As for methods for the meta embedding \cite{yin2016learning}, they concernc conducting a complementary combination of information from an ensemble of distinct word embedding sets, each trained using different methods, and resources, to yield an embedding set with improved overall quality \cite{kiela-etal-2018-dynamic, neill2018angular, coates2018frustratingly, muromagi2017linear, artetxe2018uncovering}. Thus we believe this is one of the benefits of applying Duo.

There have also been exhausted studies on the refinement of the architecture of the Transformer. Adversarial training has been proved to be beneficial to language modelling \cite{wang2019improving} . Additionally, to deal with the fix-length problem, \cite{dai2019transformer} extend the vanilla Transformer with recurrent units, which greatly enhances the original model. The Duo mechanism is a meta-embedding way to approach the Transformer.

The Duo is the first meta embedding language architecture based on attention to the best of our knowledge. In the following sections, we will describe the Duo text classification model in section \ref{subsec:Duo Classifier}, and then we will explore its implementation on encoder-decoder architecture in section \ref{subsec:duotransformer}.

\section{Model Architecture}
\subsection{Duo Classifier}
\label{subsec:Duo Classifier}

\begin{figure}[t]
\includegraphics[scale=0.4]{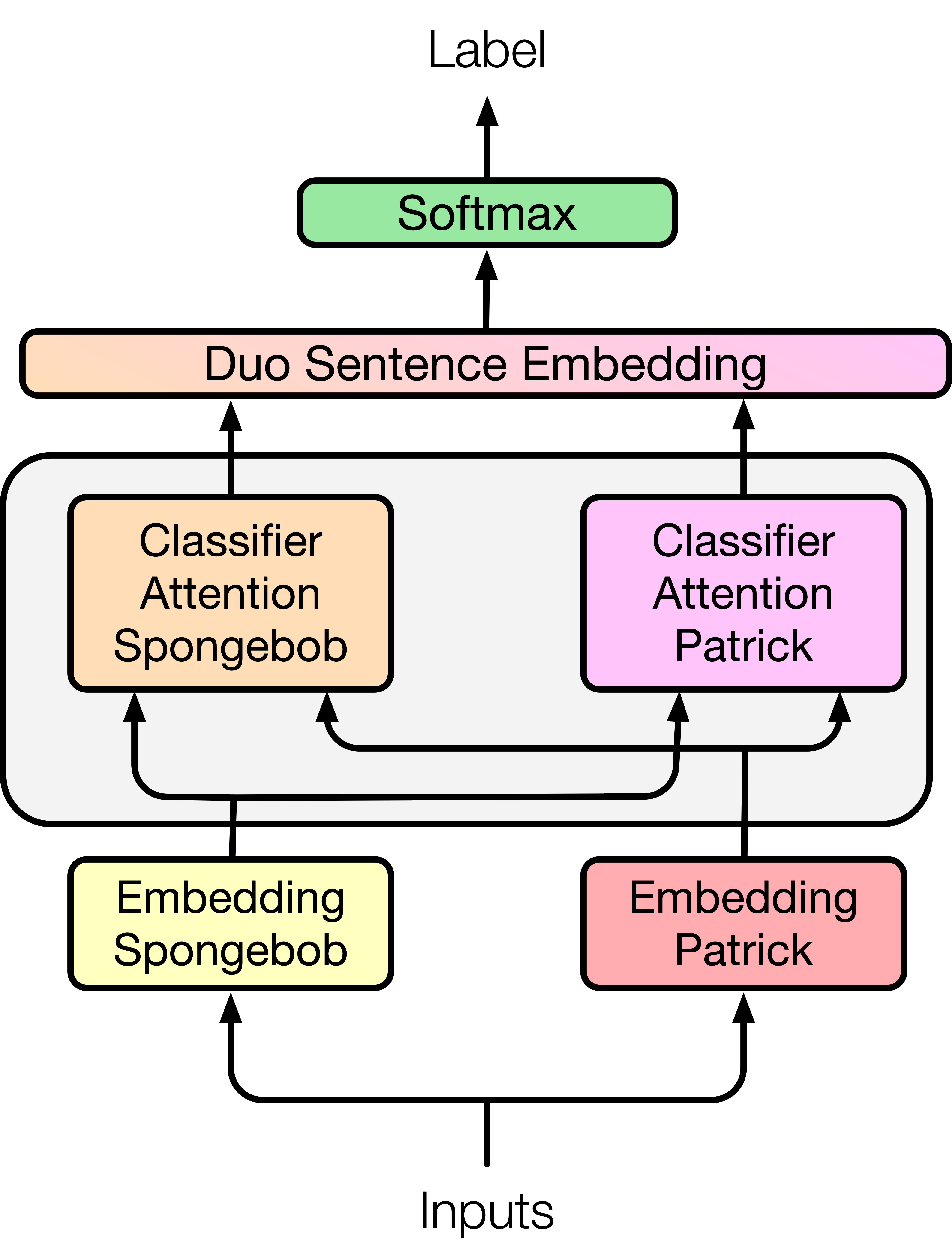}
\centering
\caption{The Duo Classifier Model Architecture}
\label{fig:classifier}
\end{figure}

\subsubsection{Duo Word Embedding}
As is demonstrated in figure  \ref{fig:classifier}, we use different embedding namely, Spongebob and Patrick, to represent the word embedding of the same input sequence $\boldsymbol{x}_i = (x_1. x_2,...,x_{l_i})$, where $l_i$ is the unfixed length of the $i$-th sentence. Later, we use embedding Spongebob and Patrick to represent separately trained word embedding, e.g., Spongebob can be GloVe 300d, and Patrick can be Word2Vec 30d. For the simply of notation, we use $S_i\in \mathbb{R}^{l_i \times d_{model1}}, P_i\in \mathbb{R}^{l_i \times d_{model2}}$ to represent the text with different word embedding.

\subsubsection{Duo Classifier Attention}
We will simply let the model to learn the parameter of attention to balance the weight of different dimension in our text classifier. In practice, we initialize two parameter $\boldsymbol{w}^{S} \in \mathbb{R}^{d_{model1}}$, and $\boldsymbol{w}^{P} \in \mathbb{R}^{d_{model2}}$.

\begin{equation}
    \boldsymbol{a_i}^P = \mbox{Attention}(\boldsymbol{w}^{S}, S_i, P_i) = \mbox{Softmax}(\boldsymbol{w}^{S}{S_i}^T)P_i
\end{equation}

\begin{equation}
    \boldsymbol{a_i}^S = \mbox{Attention}(\boldsymbol{w}^{P}, P_i, S_i) = \mbox{Softmax}(\boldsymbol{w}^{P}{P_i}^T)S_i
\end{equation}

In our later experiment, we will drop the softmax function, because doing this will have even faster computation while maintaining satisfying results.

\subsubsection{Duo Sentence Embedding}
The duo sentence embedding would be a fusion layer of $\boldsymbol{a}^{p}$ and $\boldsymbol{a}^{S}$, so, we will introduce another fusion parameter $W^O \in \mathbb{R}^{(d_{model1} + d_{model2}) \times d_{ff}}$.

\begin{equation}
    \boldsymbol{e_i} = [\boldsymbol{a_i}^P, \boldsymbol{a_i}^S]W^O
\end{equation}

where $[\cdot, \cdot]$ is concatenate operation.

The $\boldsymbol{e}$ is the final representation of sentence embedding. Its value is the weighted sum of $P, S$ based on the attention of each other. In other words, we learn the attention and value separately by giving them different embedding. 

\subsubsection{Model Complexity}
The number of parameters to be learned in our model is $d_{model_1} + d_{model_2}$ in Duo Classifier Attention layer, $(d_{model1} + d_{model2}) \times d_{ff}$ in Sentence Duo Embedding layer and $d_{ff} \times label\_num$ in the final softmax layer. If we set $d_{model_1} = d_{model_2} = 300$, $d_{ff} = 600$ and number of label = 20. The number summed up is no more than 0.4M parameters. When running on a machine with 8 GPUs, we can achieve a state-of-the result on text classification tasks 20NG in less than half an hour.

\subsection{Duo Transformer}
\label{subsec:duotransformer}

\subsubsection{Duo Attention}
After reviewing on the duo classifier, the Duo multi-head attention seems simple and straight-forward.

We have multi-head attention:

\begin{equation}
    A^S = \mbox{MultiHead}(Q^{P}, K^{P}, V^{S})
\end{equation}

\begin{equation}
    A^P = \mbox{MultiHead}(Q^{S}, K^{S}, V^{P})
\end{equation}

We can use similar formulation to calculate the Duo multi-head attention.

\begin{figure}[t]
\includegraphics[scale=0.4]{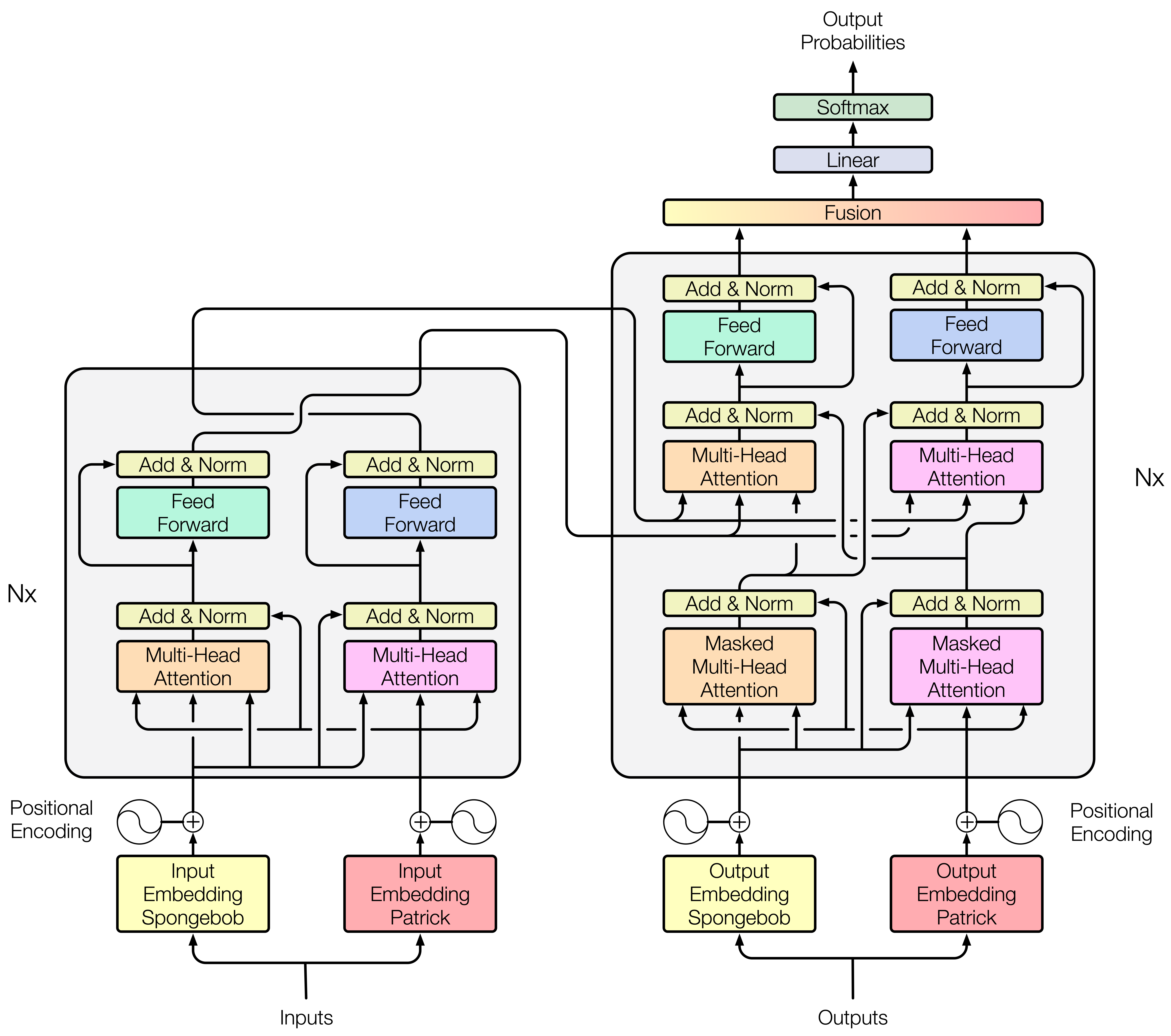}
\centering
\caption{The Duo Transformer Model Architecture}
\label{fig:transformer}
\end{figure}

From figure \ref{fig:transformer}, it seems that the number of parameters has doubled compared to vanilla attention. In order to eschew this overcomplexity, we share weight in multi-head attention of each layer. Specifically, $K^P$ and $V^p$, and $V^S$ and $K^S$ in each layer share the same projection parameters. So in the final multi-head attention, we only have $\frac{1}{3}$ more projection parameters, and our experiments show that the weight sharing result in faster convergence.

\subsection{Duo Decoder}
The Duo Decoder is quite similar to the original Transformer decoder, except the fact that the original Transformer has the same $K$ and $V$, while the Duo Decoder has different ones. We interpret that each $K$ and $V$ encode different information from each word embedding. Thus they need to be decoded separately. 

The vanilla Transformer has the same weights matrix between the two embedding layers and pre-softmax linear transformation similar to  \cite{press2016using}. However, as we have a fusion layer, we still share weights, but after a linear projection of the original concatenated duo embedding layer, and the parameters of this projection is to be learnt in the training step.

\subsection{Duo Layer Normalization}
Another intriguing part is the Duo Layer Normalization. The output of the traditional layer normalization \cite{ba2016layer} and residual connection \cite{he2016deep} is LayerNorm($x$ + Sublayer($x$)) in each unit. However, considering the dimensional difference in different word embedding, meanwhile guaranteeing more fluid cross information flow. We modify the original LayerNrom to the following formula:

\begin{equation}
    \mbox{DuoLayerNorm}(\boldsymbol{x}^S, \boldsymbol{x}^P) = \mbox{LayerNorm}(\boldsymbol{x}^S + \mbox{Sublayer}(\boldsymbol{x}^P))
\end{equation}

This mechanism is used in the decoder layer between the masked multi-head attention, and the feed-forward unit demonstrated in figure \ref{fig:transformer}.

\subsection{Difference That Matters}
Attention is undoubtedly a good idea in natural language processing. Because semantic information should be expressed via multiple dimensions, we have multi-head attention to deal with the problem, where the embedding is linearly transformed and then fed into the scaled dot-product attention. However, such a linear transformation may not contain as much information; after all, it is a simple transformation based on the same value, meaning there are still unbreakable constrains in the attention value and word value. 

Therefore with the help of meta-embeddings, the attention calculated is dependent with the word value, and with the help of Duo mechanism, the discrepancy could be efficiently narrowed down in other to serve each other from different perspective, e.g., word embedding with linear structure GloVe v.s. non-linear structure like Word2Vec.

Loosening the attention-value constrains enhances model expression ability, therefore resulting in a better performance.

\section{Experiment}
In this section, we will first demonstrate the performance of our Duo Classifier on public text classification tasks. Then we will show the results of running our model on machine translation tasks. We ran our models on 8 NVIDIA RTX 2080 Ti GPUs.

\subsection{Duo Classifier}
We compare our model with multiple state-of-the-art baselines on many public datasets in terms of accuracy. We use GloVe 50d and GloVe 300d as pre-trained embedding, which we find are the best duo couple. Then, we will run a series of self-compare experiment on different combinations of word embedding.

\subsubsection{Results}
\textbf{Settings}

We explored a variety of duo couple, and it turns out the GloVe 50d and GloVe 300d can yield the best results. Other parameters including dropout, learning rate are the same as the original transformer. We randomly selected 10\% of the training set a validation. We trained our model for a maximum of 200 epochs using Adam \cite{kingma2014adam} and stop if the validation loss does not decrease for ten consecutive epochs. The results of other models on the same datasets are from \cite{yao2019graph}. We run out models for ten times and calculate its mean. We will then further explore the result of different combinations of duo couple.

\textbf{Datasets}

We ran our experiments on five popular benchmark corpora including 20-Newsgroups (20NG)\footnote{ http://qwone.com/ ~jason/20Newsgroups/},Ohsumed\footnote{http://disi.unitn.it/moschitti/corpora.htm}, R52 and R8 of Reuters 21578\footnote{https://www.cs.umb.edu/ ~smimarog/textmining/datasets/} and Movie Review (MR)\footnote{ http://www.cs.cornell.edu/people/pabo/movie-review-data/}.
These datasets are widely used and recognized in recent publications, and we will skip the details of them. The readers could go to \cite{yao2019graph} for more detailed settings.

\textbf{Performance}

As it turns out, our model achieves the best results on 4 out of 5 benchmarks\ref{tab:testaccu}. It still ranks second in R8 dataset, and we think it is because  the words number in this datasets are less than the others(with only 7,688 words) that information simply from word embedding are not enough . 

The main reasons why duo model works well are obvious. Firstly, we use separately trained embedding. And previous research has shown that this meta embedding technology can greatly improve the performance. Secondly, we use Transformer to combine these embedding and this model is proved to be more efficient than traditional RNN-based models. Let alone we simply calculate the average of the word embedding in each documents for text classification in way of meta embedding way.
\begin{savenotes}
\begin{table}[]
\centering
\begin{tabular}{c|c|c|c|c|c}
\hline
\textbf{\textbf{Model}}                                                                                                                                  & \textbf{\textbf{20NG}}     & \textbf{\textbf{R8}}       & \textbf{\textbf{R52}}      & \textbf{\textbf{Ohsumed}}  & \textbf{\textbf{MR}}      \\ \hline
\textbf{TF-IDF+LR}\footnote{bag-of-words model with term frequency-inverse document frequency weighting. Logistic Regression is used as the classifier.} & 83.19                      & 93.74                      & 86.95                      & 54.66                      & 74.59                     \\
\textbf{CNN-rand} \cite{kim2014convolutional}                                                                                                            & 76.93                      & 94.02                      & 85.37                      & 43.87                      & 74.98                     \\
\textbf{CNN-non-static}\footnote{CNN-non-static uses pre-trained word embeddings}                                                                        & 82.15                      & 95.71                      & 87.59                      & 58.44                      & 77.75                     \\
\textbf{LSTM} \cite{liu2016recurrent}                                                                                                                    & 65.71                      & 93.68                      & 85.54                      & 41.13                      & 75.06                     \\
\textbf{LSTM}(pretrain)                                                                                                                                  & 75.43                      & 96.09                      & 90.48                      & 51.10                      & 77.33                     \\
\textbf{Bi-LSTM}                                                                                                                                         & 73.18                      & 96.31                      & 90.54                      & 49.27                      & 77.68                     \\
\textbf{PV-DBOW} \cite{le2014distributed}                                                                                                                & 74.36                      & 85.87                      & 78.29                      & 46.65                      & 61.09                     \\
\textbf{PV-DM} \cite{le2014distributed}                                                                                                                  & 51.14                      & 52.07                      & 44.92                      & 29.50                      & 59.47                     \\
\textbf{PTE} \cite{tang2015pte}                                                                                                                          & 76.74                      & 96.69                      & 90.71                      & 53.58                      & 70.23                     \\
\textbf{fastTEXT} \cite{joulin2016bag}                                                                                                                   & 79.38                      & 96.13                      & 92.81                      & 57.70                      & 75.14                     \\
\textbf{fastTEXT}(bigrams)                                                                                                                               & 79.67                      & 94.74                      & 90.99                      & 55.69                      & 76.24                     \\
\textbf{SWEM} \cite{shen2018baseline}                                                                                                                    & 85.16                      & 95.32                      & 92.94                      & 63.12                      & 76.65                     \\
\textbf{LEAM} \cite{wang2018joint}                                                                                                                       & 81.91                      & 93.31                      & 91.84                      & 58.58                      & 76.95                     \\
\textbf{Graph-CNN-C} \cite{defferrard2016convolutional}                                                                                                  & 81.42                      & 96.99                      & 92.75                      & 63.86                      & 77.22                     \\
\textbf{Graph-CNN-S} \cite{bruna2013spectral}                                                                                                            & -                          & 96.80                      & 92.74                      & 62.82                      & 76.99                     \\
\textbf{Graph-CNN-F} \cite{he2016deep}                                                                                                                   & -                          & 96.89                      & 93.20                      & 63.04                      & 76.74                     \\
\textbf{Text GCN} \cite{yao2019graph}                                                                                                                    & 86.34                      & \textbf{97.07}             & 93.56                      & 68.36                      & 76.74                     \\
\textbf{Transformer} \cite{vaswani2017attention}                                                                                                                 & -                          & 94.99                      & 86.72                      & 36.45                      & 74.85                     \\ \hline
\textbf{Ours}                                                                                                                                            & \multicolumn{1}{l|}{89.91} & \multicolumn{1}{l|}{97.02} & \multicolumn{1}{l|}{93.81} & \multicolumn{1}{l|}{71.03} & \multicolumn{1}{l}{81.02}
\end{tabular}
\end{table}
\end{savenotes}

\textbf{Which Couple Is The Best}

We explored various couples of word embedding on datasets. Including different dimensions of embedding from GloVe \cite{pennington2014glove}, CBOW \cite{mikolov2013distributed}, and fastText \cite{joulin2016bag}, and the results are demonstrated on \ref{tab:couple1}, \ref{tab:couple2} and 
, and \ref{tab:couple3}. And it turns out that the GloVe 50d and GloVe 300d duo win the competition. The result is obtained by running 10 times of different couples and calculating their mean performance on 20NG, Ohsumed and MR dataset. Without any exceptions, the Duo couple of GloVe 50d and GloVe 300d has the best results on all the tasks. These results further proves the advantages of GloVe word embedding. Additionally, it is no surprise to us that the the diagnose of the table shows relatively less satisfying results, . Because the duo embedding employs the same embedding, they are simple one-layer single-head transformer models. 

\begin{table}[]
\centering
\begin{tabular}{c|ccccc}
Word Embedding & GloVe 50d & GloVe 300d     & fastTEXT 300d & CBOW 50d & CBOW 300d \\ \hline
GloVe 50d      & 86.39     & \textbf{89.91} & 82.86         & 89.38    & 89.88     \\
GloVe 300d     & -         & 88.80          & 85.32         & 88.83    & 89.12     \\
fastTEXT 300d  & -         & -              & 81.21         & 85.23    & 82.09     \\
CBOW 50d       & -         & -              & -             & 86.69    & 79.38     \\
CBOW 300d      & -         & -              & -             & -        & 78.99    
\end{tabular}
\caption{Performance of Different  Couples on 20NG}
\label{tab:couple1}
\end{table}

\begin{table}[!ht]
\centering
\begin{tabular}{c|ccccc}
Word Embedding & GloVe 50d & GloVe 300d     & fastTEXT 300d & CBOW 50d & CBOW 300d \\ \hline
GloVe 50d      & 65.81     & \textbf{71.03} & 63.89         & 67.13    & 67.73     \\
GloVe 300d     & -         & 70.09          & 65.83         & 69.45    & 67.85     \\
fastTEXT 300d  & -         & -              & 63.09         & 61.77    & 64.35     \\
CBOW 50d       & -         & -              & -             & 62.38    & 65.49     \\
CBOW 300d      & -         & -              & -             & -        & 61.21   
\end{tabular}
\caption{Performance of Different  Couples on Ohsumed}
\label{tab:couple2}
\end{table}

\begin{table}[]
\centering
\begin{tabular}{c|ccccc}
Word Embedding & GloVe 50d & GloVe 300d     & fastTEXT 300d & CBOW 50d & CBOW 300d \\ \hline
GloVe 50d      & 80.33     & \textbf{81.02} & 78.66         & 79.67    & 69.84     \\
GloVe 300d     & -         & 76.12          & 76.42         & 73.50    & 73.94    \\
fastTEXT 300d  & -         & -              & 75.85         & 69.13    & 72.35     \\
CBOW 50d       & -         & -              & -             & 72.19    & 74.31     \\
CBOW 300d      & -         & -              & -             & -        & 73.65   
\end{tabular}
\caption{Performance of Different  Couples on MR}
\label{tab:couple3}
\end{table}

\subsection{Duo Machine Translation}

After exploring the performance of Duo in the text classification task, we further investigate whether this meta-embedding mechanism could be applied to the machine translation tasks. The potential of the model is considerable, as a good performance in task classification tasks means such a mechanism could encode a sentence much better. However, the real difficulties lay in the design of the decoder. We finally figure out a meta embedding decoder architecture based on the backbone of the Transformer demonstrated in  \ref{subsec:duotransformer}. In this part, we will examine the Duo Translator in terms of its BLEU score, and its convergence speed. 

\subsubsection{Results}

\textbf{Settings}

For the machine translation models, we followed the same hyper-parameter setup described in \cite{vaswani2017attention}. Specifically, we set the $d_{model}=512$, and the $d_{ff}$ was set to 2048. The number of layers for the encoder and the decoder was set to 8. Additionally, We use weight sharing in the Duo Multi-head to decrease the model complexity. Worth mentioning, we use gloVe 300d word embedding followed by a $300 \cdot 512$ feed-forward neural networks to fix the discrepancy of dimensionality. 

\textbf{Datasets}

On the machine translation task, we report results on three mainstream benchmark datasets: WMT 2014 English to German (En-De) consisting of about 4.5 million sentence pairs, and WMT 2014 English to French (En-Fr) of 36M sentences. We used byte-pair encoding \cite{britz2017massive} of size 32K and 40K tokens for each task.

\textbf{Performance}

We demonstrate the effectiveness of our model in table \ref{tab:mtaccu}, which shows that meta embedding could clearly benefit the process of translation. Specifically, our model is able to achieve a state-of-the-art score on WMT 2014 En-De benchmark, and still competitive in WMT 2014 English to France bench mark. Worth mentioning, the meta-embedding Duo Transformer has outperformed the vanilla transformer by 1.3 and 1.1 BLEU score on each task, further proving the advantage of the meta-embedding mechanism.

\begin{table}[]
\centering
\begin{tabular}{c|ccc}
\hline
\textbf{Model}                                & \textbf{Param} & \textbf{WMT En-De} & \textbf{WMT En-Fr} \\ \hline
ConvS2S \cite{gehring2017convolutional}       & 216M                  & 25.2               & 40.5               \\
Transformer big \cite{vaswani2017attention}       & 213M                  & 28.4               & 41.0               \\
Weighted Transformer \cite{ahmed2017weighted} & 213M                  & 28.9               & 41.4               \\
RNMT+ \cite{chen2018best}                     & 379M                  & 28.5               & 41.0               \\
Transformer with RPP \cite{shaw2018self}      & -                     & 29.2               & 41.5               \\
SNMT \cite{ott2018scaling}                    & 210M                  & 29.3               & \textbf{43.2}      \\
DynamicConv \cite{wu2019pay}                  & 213M                  & \textbf{29.7}      & \textbf{43.2}      \\
TaLK Convolution \cite{lioutas2020time}       & 209M                   & 29.6               & \textbf{43.2}      \\ \hline
\textbf{Ours}                                 & 220M                  & \textbf{29.7}      & 42.1              
\end{tabular}
\caption{Machine translation accuracy in terms of BLEU for WMT En-De and WMT En-Fr on newstest2014.}
\label{tab:mtaccu}
\end{table}

\begin{figure}[t]
\includegraphics[scale=0.5]{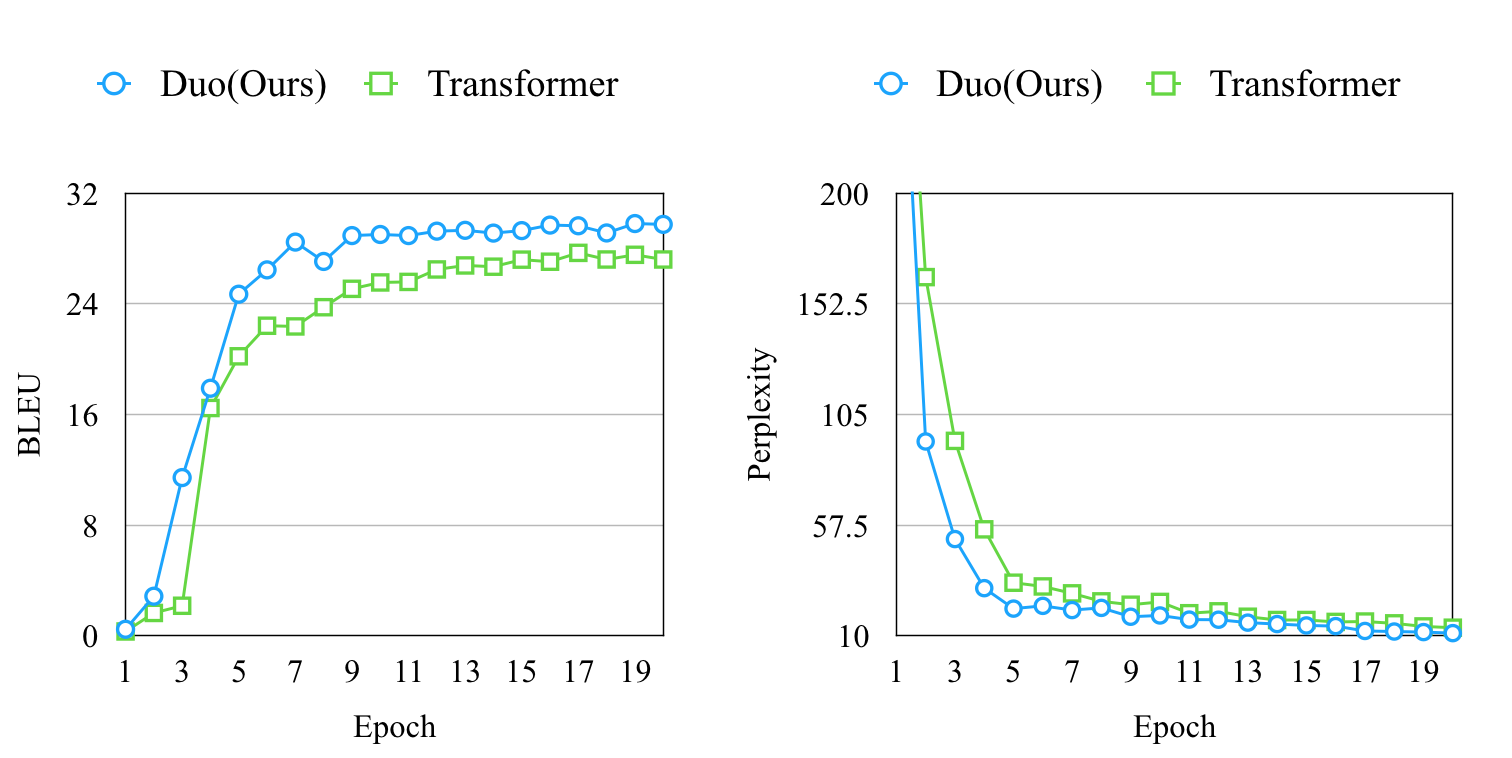}
\centering
\caption{BLEU and Perplexity on WMT 2014 En-De Validation Set}
\label{fig:trainprocess}
\end{figure}

Figure \ref{fig:trainprocess}, alongside with table \ref{table:perdetail} and \ref{table:bleudetail} also demonstrates the faster convergence, as well as better performance results by meta embeddings. The results are obtained by calculating the average of 3 separate runnings of each model on the WMT 2014 En-De Validation Set.

\begin{table}[t]
\centering
\begin{tabular}{lll}
\hline
\multicolumn{1}{c}{Model}         & \multicolumn{1}{c}{Param} & \multicolumn{1}{c}{BLEU}  \\ \hline
Transformer    & \multicolumn{1}{c}{213M}  & \multicolumn{1}{c}{$27.31 \pm 0.01$} \\
+ Meta Embeddings                 & \multicolumn{1}{c}{246M}  & \multicolumn{1}{c}{$28.41\pm0.2$} \\
+ Weight Sharing in Duo Multihead & 220M                      & $28.58\pm0.05$                     \\
+ Duo Normalization               & 220M                      & $29.60\pm0.07$                     \\
+ Fusion Layer                    & 220M                      & $29.68\pm0.03$                     \\ \hline
\end{tabular}
\caption{Ablation on WMT En-De validation set. (+) indicates that a result includes all preceding features.}
\label{table:abla}
\end{table}

In order to evaluate the function of different parts in our architecture, we did an ablation test on the WMT 2014 En-De Validation set. We used the same hyper-parameters as before, and the results are reported in Table \ref{table:abla}. Initially, we add the meta embeddings to the Vanilla Transformer Model, and it seems that this gives the most salient advancement of the performance. However, the number of parameters increased quite a lot, and the improvement may merely come from the additional parameter numbers. Therefore, we decided to shrink the model's size by using weight sharing in Duo Multihead. It turns out that this operation not only reduced the number of parameters but also improves the performance. The following Normalization and Fusion layer has also been proved to be beneficial.

\begin{table}[!h]
\centering
\begin{tabular}{c|ccccccccccccc}
\hline
\# Epoch      & ...      & 10    & 11    & 12    & 13    & 14    & 15   & 16    & 17    & 18    & 19    & 20    \\ \hline
Transformer & ... & 25.55 & 25.59 & 26.49 & 26.78 & 26.69 & 27.2 & 27.05 & 27.69 & 27.21 & 27.55 & 27.21 \\
\textbf{Duo(Ours)}   & ...  & 29.02 & 28.94 & 29.26 & 29.32 & 29.12 & 29.3 & 29.7  & 29.65 & 29.13 & 29.69 & 29.75
\end{tabular}
\caption{Details of BLEU on WMT 2014 En-De Validation Set for the last 11 epoches}
\label{table:bleudetail}
\begin{tabular}{c|ccccccccccccc}
\hline
\# Epoch       & ...      & 10    & 11    & 12    & 13    & 14    & 15    & 16    & 17    & 18    & 19    & 20    \\ \hline
Transformer & ...  & 24.53 & 19.68 & 20.49 & 18.17 & 16.77 & 16.77 & 15.95 & 16.11 & 15.33 & 14.01 & 13.46 \\
\textbf{Duo(Ours)}   & ...  & 18.72 & 16.94 & 16.89 & 15.64 & 15.03 & 14.43 & 14.15 & 12.06 & 11.82 & 11.58 & 11.13
\end{tabular}
\caption{Details of Perlexity on WMT 2014 En-De Validation Set for the last 11 epoches}
\label{table:perdetail}
\end{table}

\section{Conclusion}

In this work, we presented the Duo Model, the first meta-embeddings mechanism based on self-attention, which improves the performance of language modelling by exploiting more than one word-embedding.

For text-classification tasks, a single-layer Duo Classifier can achieve the state-of-the-art results on many public benchmarks. Moreover, for machine translation tasks, we introduce the first encoder-decoder models with more than one embedding. Furthermore, we prove that this meta embedding mechanism benefits the vanilla transformer in terms of not only better performance but also a faster convergence.

Nowadays, though there is more and more attention paid to meta-mebeddings in natural language   processing, we still think that this mechanism has potential other than the text classification task. We sincerely expect more investigations into this field.

\bibliographystyle{acm}

\bibliography{template}  %%% Remove comment to use the external .bib file (using bibtex).

%\begin{thebibliography}{1}
%
%\bibitem{kour2014real}
%George Kour and Raid Saabne.
%\newblock Real-time segmentation of on-line handwritten arabic %script.
%\newblock In {\em Frontiers in Handwriting Recognition (ICFHR), %2014 14th
%  International Conference on}, pages 417--422. IEEE, 2014.
%
%\bibitem{kour2014fast}
%George Kour and Raid Saabne.
%\newblock Fast classification of handwritten on-line arabic %characters.
%\newblock In {\em Soft Computing and Pattern Recognition %(SoCPaR), 2014 6th
%  International Conference of}, pages 312--318. IEEE, 2014.
%
%\bibitem{hadash2018estimate}
%Guy Hadash, Einat Kermany, Boaz Carmeli, Ofer Lavi, George Kour, %and Alon
%  Jacovi.
%\newblock Estimate and replace: A novel approach to integrating %deep neural
%  networks with existing applications.
%\newblock {\em arXiv preprint arXiv:1804.09028}, 2018.
%
%\end{thebibliography}

\end{document}